\documentclass[11pt]{article}
\usepackage[ruled,vlined,linesnumbered]{algorithm2e}

\usepackage[preprint]{acl}

\usepackage{times}
\usepackage{latexsym}
\usepackage{amsmath}
\usepackage{booktabs}
\usepackage{multirow}
\usepackage{arydshln}

\usepackage[T1]{fontenc}

\usepackage[utf8]{inputenc}

\usepackage{microtype}

\usepackage{inconsolata}

\usepackage{graphicx}

\title{Words Are the Voice of the Mind: Depression Severity Estimation from Social Traces via Mental-Guided Evidence Reasoning}

\author{
Nada Xu\textsuperscript{1},
Xueqin Chen\textsuperscript{1},
Qiang Gao\textsuperscript{2},
Hongru Liang\textsuperscript{1},
Wenqiang Lei\textsuperscript{1} \\
\textsuperscript{1}Sichuan University \\
\textsuperscript{2}Southwestern University of Finance and Economics \\
\texttt{xunada@stu.scu.edu.cn}
}

\definecolor{textpurple}{RGB}{0,0,0}
\definecolor{myblue}{RGB}{0, 0, 255}
\definecolor{mygray}{gray}{0.6}
\definecolor{lightgray}{gray}{0.91}

\definecolor{upcolor}{RGB}{57,182,74}

\definecolor{goodblue}{HTML}{0071bc}
\hypersetup{colorlinks=true,breaklinks,linkcolor=red,citecolor=goodblue}

 \AtEndPreamble{
    \usepackage[capitalize]{cleveref}
    \crefname{section}{Sec.}{Secs.}
    \Crefname{section}{Section}{Sections}
    \crefname{table}{Tab.}{Tabs.}
    \Crefname{table}{Table}{Tables}
}

\newcommand{\task}{ST-DSE}
\newcommand{\method}{UMer}
\newcommand{\HumanMark}{\textsuperscript{\ensuremath{\dagger}}}
\newcommand{\TrainMark}{\textsuperscript{\ensuremath{\ddagger}}}

\begin{document}
\maketitle

\begin{abstract}
Depression severity estimation is crucial for mental health assessment and intervention, yet traditional approaches depend on clinical interviews or self-report questionnaires that require users' active participation. 
With the widespread use of online social media, users leave rich social traces that may reflect their mental states, motivating the task of social media-based depression severity estimation.
Existing efforts have progressed from textual-feature-based classification to LLM-based questionnaire completion over standardized scale items, yet they still fall short in user-specific mental priors modeling, diversified semantic retrieval, and fine-grained reasoning over mental evidence.
To address these limitations, we propose \method, a user-specific mental evidence reasoning framework for social traces-based depression severity estimation. \method~first distills a user-specific mental prior from historical social traces to capture the user's psychological tendencies, behavioral patterns, and self-expression style. 
It then performs scale-guided multi-query retrieval to construct evidence memories from item-relevant posts.
Finally, it conducts coarse-to-fine evidence-aware reasoning to distinguish direct and indirect evidence and predict item-level scores with interpretable rationales. 
Extensive experiments on real-world benchmarks demonstrate that \method~not only achieves competitive performance over existing baselines, but also provides evidence-grounded interpretability and generalizes to other scale-based tasks.
\end{abstract}

\section{Introduction}
Depression has become a growing public health concern in modern society, due to individuals being increasingly exposed to stress from work, study, relationships, and daily life~\cite{magomedova2025mental}. Traditional depression assessment methods, such as clinical interviews and standardized self-report questionnaires, are effective but require that users actively seek help~\cite{joiner2005evidence}. 
Recently, online social media has provided a spontaneous space for people to communicate, express themselves, and share daily experiences, where users leave rich social traces, such as posts, comments, replies, and interaction histories, that may reveal their personality traits, behavioral patterns, and psychological states that are difficult to capture through traditional offline assessments, making them a promising source for depression severity estimation~\cite{masood2019ude}.

\begin{figure}[htb]
\vspace{-2mm}
    \centering
    \includegraphics[width=0.49\textwidth]{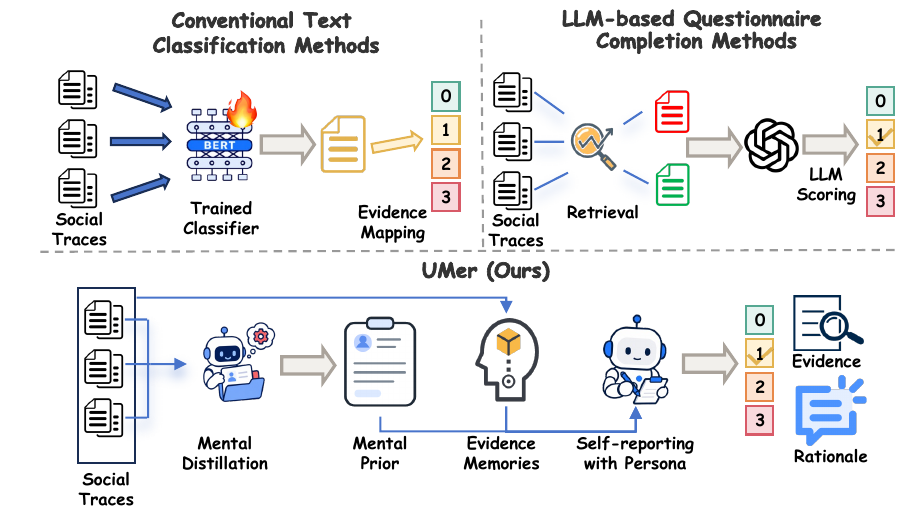}
    \vspace{-6mm}
    \caption{Existing Efforts vs. \method.}
    \label{fig:intro}
    \vspace{-2mm}

\end{figure}


Thus, researchers have reformulated depression assessment as a social traces-based depression severity estimation (\task) task, which aims to infer users' risk levels from their social media posts~\cite{losada-etal-2020-erisk,parapar2021overview}. Earlier works typically treated \task~as a text classification task, relying on extracted textual features from social traces to train supervised classifiers~\citep{ji2022mentalbert}. Although these methods can achieve relatively high accuracy, they mainly predict depression severity from uninterpretable surface-level textual signals and require sufficient labeled data for effective training. 
In the era of large language models (LLMs), recent studies have begun to reformulate \task~as a questionnaire scoring problem~\citep{aragon2024delving,ravenda-etal-2025-llms}, where LLMs are prompted to assign scores to standardized scale items based on users' social media posts. This transformation provides an evidence-based solution and improves interpretability by aligning social media evidence with item-level psychological measurements.
However, the following limitations remain:

\noindent(\textbf{L1}) \textit{Overlooking the user-perspective nature of self-report assessment}. Existing LLM-based methods primarily retrieve posts as supporting evidence for scale-level prediction. However, standardized depression scales are originally designed as self-report questionnaires, where each item reflects how a specific user would assess their own mental state. Therefore, the model should not only retrieve item-specific evidence from users' social traces, but also capture the user's overall mental state, self-expression style, and behavioral patterns from these records, thereby fulfilling the task through user-specific self-report simulation.

\noindent(\textbf{L2}) \textit{Incomplete semantic coverage of evidence
}. Existing methods for evidence retrieval often rely on coarse-grained embedding similarity between standardized scale items and user posts. However, scale items are usually written in medical and psychometric language, while users describe their mental states through daily events, fragmented feelings, implicit behaviors, and colloquial expressions. Therefore, directly using the original scale item descriptions as retrieval queries may fail to retrieve posts that are psychologically relevant but lexically different~\citep{zeng2006assisting}, resulting in incomplete semantic coverage of evidence.

\noindent(\textbf{L3}) \textit{Insufficient fine-grained evidence reasoning}. Current LLM-based methods make predictions directly based on retrieved posts, without explicitly assessing the role and strength of each piece of evidence. However, retrieved posts may support a scale item in different ways. For example, some posts directly match the item definition through explicit symptom descriptions or clear keyword-level expressions, while others only provide indirect evidence through daily-life events, behavioral changes, contextual self-disclosures, or long-term expression patterns. Such evidence requires structured reasoning over relevance, evidential strength, and user-specific context, rather than simply relying on semantic similarity or keyword overlap.

To address these limitations, we propose a \underline{U}ser-specific \underline{M}ental-guided \underline{E}vidence \underline{R}easoning framework (\method). 
Specifically, \method~is built upon LLM in an agentic paradigm, which follows a three-stage workflow. \method~first distills a user-specific mental prior from the user's historical social traces to understand the user's psychological tendencies, behavioral patterns, and self-expression style. 
It then performs scale-guided multi-query retrieval to obtain item-related posts from multiple semantic perspectives, thereby forming scale-aware mental evidence memories for each user. Finally, a coarse-to-fine evidence-aware reasoning mechanism is utilized to guide the LLM to distinguish direct and indirect evidence and predict item-level scores with evidence-grounded rationales. We conduct experiments on real-world benchmark datasets, demonstrating not only the effectiveness and interpretability of \method~compared with existing baselines, but also its generalization ability across different mental health computing scenarios. In sum, our main contributions are summarized as follows:

\begin{itemize}
    \item We reformulate \task~as a user-specific self-report inference task, emphasizing that LLMs should reason from the user's perspective rather than simply mapping retrieved social traces to a specific scale item. To support this, we distill a user-specific mental prior from historical social traces to reflect the user's inner mental characteristics, enabling the LLM to perform user-perspective mental reasoning.

    \item We propose \method, a mental-guided evidence-aware reasoning framework for \task. \method~is equipped with mental prior and mental evidence memories constructed through scale-guided multi-query retrieval. It further performs coarse-to-fine reasoning, distinguishing direct from indirect evidence for interpretable item-level scoring.

    \item We conduct extensive experiments on real-world benchmarks, demonstrating that \method~outperforms existing baselines in depression severity estimation while providing evidence-grounded rationales and showing generalization ability across different mental health computing scenarios.

\end{itemize}

\section{Related Work}
Users' social traces in social media have increasingly been used to model mental well-being and understand health-related outcomes~\cite{chancellor2020methods}. Existing studies on \task~mainly fall into the following two paradigms: 

\noindent\textbf{Conventional Text Classification Methods}. Conventional works focus on extracting lexical or semantic features from users' social media histories and train supervised classifiers by treating severity scores as classification labels. For instance, \citet{uban-rosso-2020-deep} combine psycholinguistic, emotion, and sentence-embedding features with SVM and logistic regression classifiers for depression-level prediction. 
Subsequently, the rise of pre-trained language models shifted the field toward contextual semantic modeling, which applies BERT-based classifiers to infer depression severity from users' social media histories~\citep{martinez-castano-etal-2020-early}.
The deployment of domain-adaptive models, such as MentalBERT and MentalRoBERTa, further shows that continued pre-training on mental-health-related corpora can improve mental health risk detection~\citep{ji2022mentalbert}. In addition, \citet{perez2022automatic} develops symptom-specific word embedding models to estimate BDI-II scores, while \citet{perez-etal-2023-semantic} proposes semantic similarity models to retrieve posts relevant to depression severity levels, providing interpretable evidence for severity estimation. Although these methods can achieve competitive accuracy, their reliance on supervised training may limit generalization ability and reduce interpretability.

\noindent\textbf{LLM-based Questionnaire Completion Approaches}.
Recent studies have explored the potential of LLMs in mental health analysis~\citep{aragon2024delving}, emotional state understanding, and explanation generation from social media posts~\citep{yang2023towards,xu2024mentalllm}. For instance, \citet{chim-etal-2024-overview} utilizes LLMs for suicide-risk analysis by identifying risk-indicative evidence and producing interpretable rationales, making timely online risk assessment more feasible. \citet{aragon2024delving} assesses depression severity by predicting sentence-level summaries constructed separately for each BDI item. \citet{ravenda-etal-2025-llms} propose Adaptive Retrieval-Augmented Generation (aRAG) for zero-shot psychological questionnaire completion from social media posts. \citet{ravenda2025steering} investigates steering vectors as a lightweight alternative for improving LLM-based mental health assessment. However, existing LLM-based approaches still largely rely on direct prompting or coarse retrieval over social media posts, making it difficult to fully model user-specific mental context and perform fine-grained reasoning over direct and indirect evidence. 
Our \textbf{\method}~moves beyond direct questionnaire completion by introducing a user-specific reasoning framework. It distills a mental prior and organizes scale-aware mental memories from users' historical social traces, and then performs fine-grained evidence-aware reasoning to distinguish direct and indirect evidence for interpretable item-level scoring at a user perspective.

\section{Methodology}
\begin{figure*}[t]
    \centering
    \includegraphics[width=\textwidth]{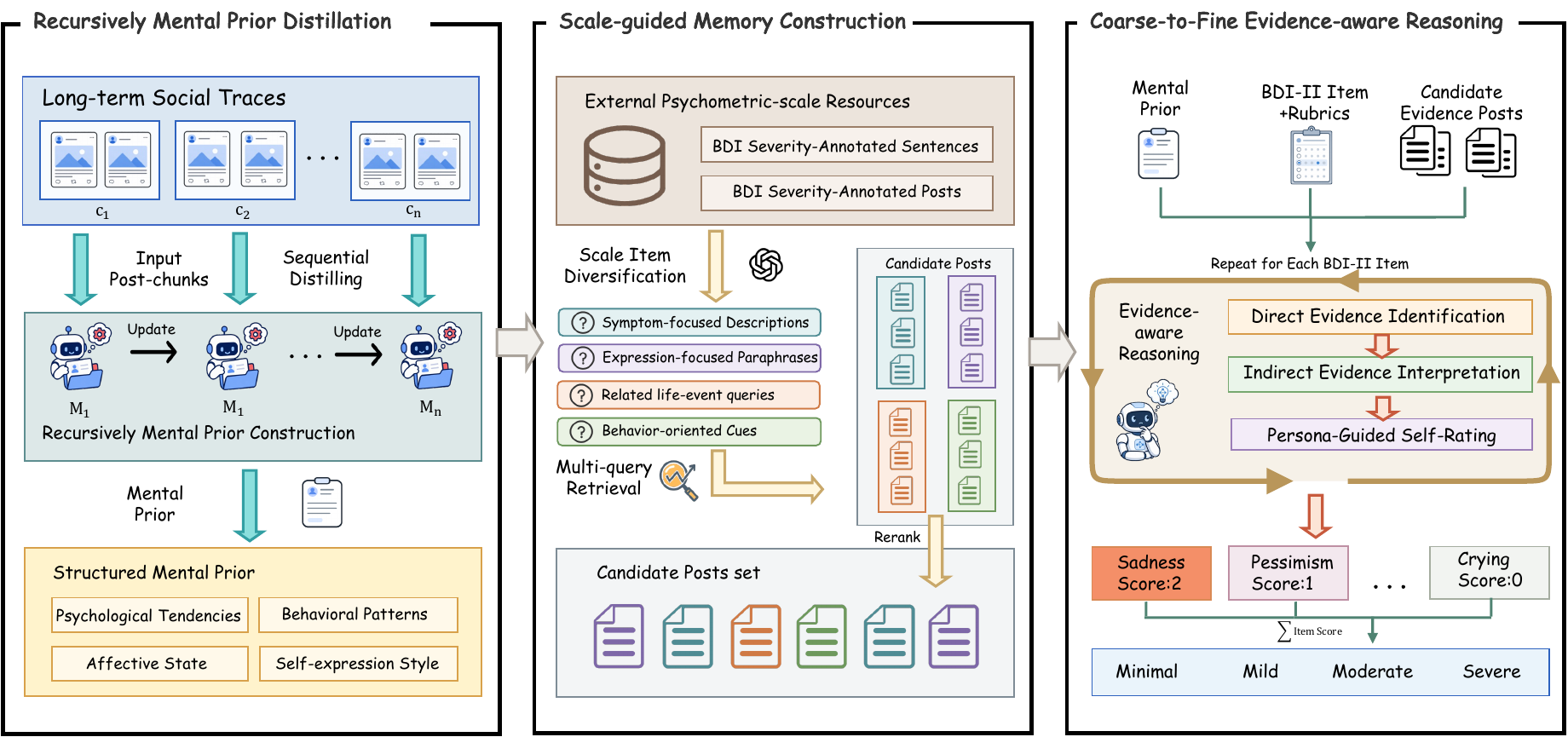}
    \caption{
    Overview of \method, a three-stage LLM-based agentic framework. The model progressively constructs the user-specific mental prior from historical social traces, retrieves item-relevant posts to construct scale-aware mental evidence memories, and performs coarse-to-fine evidence-aware reasoning for interpretable item-level scoring.
    }
    \label{fig:framework}
\end{figure*}

In this work, we focus on \textit{Social Traces-based Depression Severity Estimation} (ST-DSE), which aims to estimate a user's depression severity by retrieving scale item-relevant evidence from his or her historical social traces.
Unlike existing efforts, we formulate \task~as a self-report simulation task that infers how a user would respond to standardized scale items based on the user's historical social traces, and propose \method, a user-specific persona-guided mental reasoning framework.
As shown in Fig.~\ref{fig:framework}, \method~follows a three-stage agentic workflow: 1) \textit{mental prior distillation}, 2) \textit{scale-guided mental memory retrieval}, and 3) \textit{evidence-aware reasoning}. The details of each stage are presented in the following sections.

\subsection{Recursively Mental Prior Distillation}

Since \method~aims to simulate users' self-reporting process on standardized scale items, it requires the LLM to reason from the user's perspective. Thus, accurate mental prior are essential for providing user-level priors that help the LLM understand individual characteristics and behavioral patterns~\cite{jansen2020persona}.
To fulfill this step, we utilize an LLM to summarize the user's behavioral patterns, personality-related tendencies, and self-expression style from social traces, thereby generating a natural-language mental prior~\cite{zhang2024guided}.

Instead of directly feeding the entire history into an LLM, which not only introduces substantial computational burden and performance degradation but also makes it difficult to utilize key information in long contexts~\cite{jiang2024longllmlingua}, we design a recursive summarization mechanism~\cite{wang2025recursive_memory} to construct the user-specific mental prior through recursive chunk updating.
Specifically, given a user's long-term social traces \(X_u=\{x_1,x_2,\ldots,x_n\}\), where each \(x_i\) denotes a historical post, we first split them into a sequence of fixed-length chunks:
\begin{equation}
    C=\{c_1,c_2,\ldots,c_m\}.
\end{equation}
Subsequently, we leverage a mental prior distilling skill (see Appendix~\ref{app:ms_prompt}), designed following indirect mental prior assessment~\citep{immelman2002political}, to guide the LLM in recursively distilling the user's mental prior, thereby capturing the user's psychological tendencies, behavioral patterns, affective state, and self-expression style. This process is formulated as follows:
\begin{equation}
p(S_{1:m} \mid c_{1:m})
=
\prod_{i=1}^{m} p(S_i \mid S_{i-1}, c_i),
\end{equation}
where \(S_0\) denotes an empty initial prior, and each prior state \(S_i\) is updated from \(S_{i-1}\) using the current chunk \(c_i\). The final state \(S_m\) is used as the mental prior \(\mathcal{S}_u\).


\subsection{Scale-guided Memory Construction}
To mitigate the semantic gap between social media language and clinical criteria, unlike existing methods that retrieve evidence by directly computing the similarity between standardized scale items and user posts, we design a \textit{Scale-guided Multi-query Retrieval} (SMR) strategy. Specifically, SMR consists of two stages: 1) \textit{scale-item diversification}, which rewrites standardized scale items into post-oriented semantic probes to remedy the scale--trace semantic gap caused by the mismatch between psychometric descriptions and informal user expressions, and 2) \textit{mental evidence retrieval}, which retrieves candidate item-relevant posts and organizes them into the user's mental evidence memory for subsequent reasoning.

\noindent \textbf{Diversification.} Instead of directly using standardized scale items as retrieval queries, we first construct an item-specific rubric set for each scale item \(i_j\). Specifically, we combine its official description \(d_j\) with symptom-relevant examples from external psychometric-scale resources (e.g., DepreSym~\citep{perez2025depresym}), forming a severity-aware rubric set $\mathcal{R}_{j}=\{\mathcal{R}_{j}^{0},\mathcal{R}_{j}^{1},\mathcal{R}_{j}^{2},\mathcal{R}_{j}^{3}\}.$ Here, \(\mathcal{R}_{j}^{s}\) denotes the rubric examples corresponding to severity level \(s \in \{0,1,2,3\}\), and each entry contains realistic symptom expressions, event descriptions, behavioral cues, and lexical variants that may appear in social media posts.

Subsequently, we employ an LLM to transform the official item description \(d_j\) based on the rubric set \(\mathcal{R}_{j}\) into a multi-path query set:
\begin{equation}
    \mathcal{Q}_{j}=\text{LLM}(d_j,\mathcal{R}_{j}).
\end{equation}
The resulting queries finally cover multiple evidence-seeking views of the same symptom dimension, including symptom-focused descriptions, expression-focused paraphrases, related life-event queries, and behavior-oriented cues. These queries then serve as diversified semantic probes for item-relevant evidence retrieval.


\noindent \textbf{Retrieval.} For each query \(q \in \mathcal{Q}_{j}\), we retrieve candidate posts from the user's historical social traces \(X_u\) by selecting the top-\(K\) posts with the highest embedding similarity:
\begin{equation}
    \mathcal{T}_{u,j}^{q}
=
\operatorname{TopK}_{x \in X_u}
\cos \bigl(\phi(q), \phi(x)\bigr),
\end{equation}
where \(\phi(\cdot)\) denotes the embedding encoder, and \(\cos(\cdot,\cdot)\) denotes cosine similarity. Each \(\mathcal{T}_{u,j}^{q}\) contains candidate posts retrieved from one semantic probe of the \(j\)-th scale item.
We then merge the retrieved post sets from all diversified queries and form an item-specific mental evidence memory:
\begin{equation}
   \mathcal{M}_{u,j}
=
\operatorname{Unique}
\left(
\cup_{q \in \mathcal{Q}_{j}} \mathcal{T}_{u,j}^{q}
\right). 
\end{equation}
Here, \(\operatorname{Unique}(\cdot)\) denotes a score-based deduplication function, which is utilized to filter out duplicate posts across different query-specific candidate sets $\mathcal{T}_{u,j}^q$ by solely retaining the instance with the highest similarity score.
%
Finally, for each scale item \(i_j\), we retain only the top-\(K\) candidate evidences in \(\mathcal{M}_{u,j}\) according to their updated similarity scores:
\begin{equation}
E_{u,j}
=
\operatorname{TopK}_{x \in \mathcal{M}_{u,j}}
\operatorname{Score}(x).
\end{equation}
By repeating this process for all scale items, we construct the user's scale-aware mental evidence memory $\mathcal{E}_u=\{E_{u,1},E_{u,2},\ldots,E_{u,J}\}$.

\subsection{Coarse-to-Fine Evidence-aware Reasoning}

After the retrieval stage, we obtain a scale-aware mental evidence memory \(\mathcal{E}_u\) for each user. Existing methods typically make predictions directly based on the retrieved evidence, but they fall short in fine-grained evidence assessment and interpretable rationale generation~\citep{bao2025redsm5,perez2025depresym}. Therefore, we design a coarse-to-fine evidence-aware reasoning (EaR) mechanism to guide the LLM to assess the retrieved posts before producing the final item-level score. The EaR process is formulated as:
\begin{equation}
    D_{u,j}, I_{u,j}
=
\operatorname{Reason}_{\theta}
\left(
d_j, \mathcal{R}_{j}, E_{u,j}, \mathcal{S}_u
\right),
\end{equation}
where \(D_{u,j}\) and \(I_{u,j}\) denote the direct and indirect evidence, respectively.

Specifically, for each scale item \(i_j\), the EaR follows three steps. 
(S1) The LLM identifies direct evidence \(D_{u,j}\) from \(E_{u,j}\), where the posts explicitly describe symptoms or states that closely match the definition \(d_j\) of \(i_j\). 
(S2) Equipped with the user-specific mental prior \(\mathcal{S}_u\), the LLM reasons from the user's perspective to identify indirect evidence \(I_{u,j}\), where the posts do not explicitly mention the target symptom but may imply it through life events, behavioral changes, self-expression patterns, or contextual signals. 
(S3) The LLM integrates both direct and indirect evidence by jointly considering their relevance, evidential strength, severity-level similarity to the rubric examples \(\mathcal{R}_j\), and consistency across posts. 
In a nutshell, this process encourages the model to assess evidence quality rather than simply relying on semantic similarity or keyword overlap.

\subsection{Depression Severity Estimation}
Finally, for each scale item \(q_j\), the model predicts an item-level self-rating score:
\begin{equation}
\small
    \hat{y}_{u,j}
    =
    \arg\max_{s \in \{0,1,2,3\}}
    p_{\theta}
    \left(
    s \mid d_j, \mathcal{R}_{j}^{s}, D_{u,j}, I_{u,j}, \mathcal{S}_u
    \right),
\end{equation}
where \(\hat{y}_{u,j}\) denotes the predicted self-rating score of user \(u\) on the \(j\)-th scale item. Along with each item-level score, the model synchronously outputs the supporting evidence posts and a concise simulated self-rating. After all item scores are obtained, they are summed into a total scale score ranging from 0 to 63, which is directly mapped to the standard severity categories ( Minimal (\(0\)--\(13\)), Mild (\(14\)--\(19\)), Moderate (\(20\)--\(28\)), and Severe (\(29\)--\(63\))).

\section{Experiment}
\subsection{Experiment Settings}
\noindent\textbf{Datasets \& Metrics}. We conduct depression severity prediction experiments on the eRisk 2020~\cite{losada-etal-2020-erisk} and eRisk 2021 datasets~\cite{parapar2021erisk}. Following the official eRisk evaluation protocol~\citep{losada-etal-2020-erisk}, we report two item-level metrics, ACR and AHR, and two questionnaire-level metrics, ADODL and DCHR. In addition, we use the eRisk 2022~\citep{parapar-etal-2022-erisk} and eRisk 2023~\citep{parapar-etal-2023-erisk} datasets to evaluate \method's generalization ability. 
All eRisk data were accessed and used in accordance with the official User Agreement.

\begin{table*}[!htb]
\centering
\footnotesize
\setlength{\tabcolsep}{4pt}
\renewcommand{\arraystretch}{1.08}
\caption{Depression severity estimation on eRisk 2020 and eRisk 2021. \ensuremath{\dagger}: uses manual annotations;
\ensuremath{\ddagger}: involves task-specific model training. The best results are highlighted in \textbf{bold}, and the second-best results are \underline{underlined}.}
\vspace{-2mm}
\label{tab:main_results}
\begin{tabular*}{\textwidth}{@{\extracolsep{\fill}}llcccc@{}}
\toprule
& & \multicolumn{2}{c}{\textbf{Questionnaire Metrics}} 
& \multicolumn{2}{c}{\textbf{Symptom Metrics}} \\
\cmidrule(lr){3-4} \cmidrule(lr){5-6}
\textbf{Collection} & \textbf{Model} 
& \textbf{DCHR}~\% ($\uparrow$) 
& \textbf{ADODL}~\% ($\uparrow$) 
& \textbf{AHR}~\% ($\uparrow$) 
& \textbf{ACR}~\% ($\uparrow$) \\
\midrule

\multirow{14}{*}{\textbf{eRisk 2020}}
& BioInfo\TrainMark                                      & 30.00 & 76.01 & 38.30 & 69.21 \\
& iLab\TrainMark                                        & 27.14 & 81.70 & 37.07 & 69.41 \\
& RELAI\TrainMark                                       & 34.29 & 83.15 & 36.39 & 68.32 \\
& Sense2Vec\TrainMark         & 37.14 & 82.61 & 38.97 & 70.10 \\
& Voting\HumanMark               & 47.14 & \textbf{85.33} & 35.24 & 67.41 \\
& Recall\HumanMark               & \textbf{50.00} & \underline{85.24} & 35.44 & 67.23 \\
\cdashline{2-6}
& aRAG-CoT (GPT-5.2)                          & 25.71 & 79.89 & 40.95 & 72.13 \\
& aRAG-CoT (Gemini-3.1-Pro)                   & 22.86 & 75.10 & 40.95 & 70.48 \\
& aRAG-CoT (Kimi-K2.5)                        & 22.86 & 76.03 & 41.22 & 71.59 \\
& aRAG-CoT (Qwen3-14B)                        & 24.29 & 76.62 & 38.50 & 68.50 \\
& aRAG-CoT (DeepSeek-R1-Distill-Llama-8B)     & 35.71 & 82.04 & 33.13 & 67.85 \\
\cdashline{2-6}
& \textbf{\method} (GPT-5.2)                     & \underline{40.00} & 84.13 & \underline{41.97} & \textbf{73.77} \\
& \textbf{\method} (Gemini-3.1-Pro)              & 36.23 & 80.95 & \textbf{43.13} & 73.04 \\
& \textbf{\method} (Kimi-K2.5)                   & 35.71 & 82.79 & 41.09 & \underline{73.17} \\
& \textbf{\method} (Qwen3-14B)                   & 34.29 & 84.15 & 37.76 & 72.81 \\
& \textbf{\method} (DeepSeek-R1-Distill-Llama-8B)& 35.71 & 84.15 & 34.56 & 72.59 \\

\midrule

\multirow{13}{*}{\textbf{eRisk 2021}}
& DUTH\TrainMark                                         & 15.00 & 73.97 & \textbf{35.36} & 67.18 \\
& Symanto\TrainMark                                      & 32.50 & 82.42 & 34.17 & \textbf{73.17} \\
& CYUT\TrainMark                                         & 41.25 & \underline{83.59} & 32.62 & 69.46 \\
& Voting\HumanMark                & 46.25 & 81.72 & 27.95 & 62.40 \\
& Recall\HumanMark                & \underline{51.25} & 81.65 & 27.66 & 61.72 \\
\cdashline{2-6}
& aRAG-CoT (GPT-5.2)                          & 16.25 & 71.94 & 28.81 & 64.15 \\
& aRAG-CoT (Gemini-3.1-Pro)                   & 18.75 & 71.15 & 29.56 & 64.23 \\
& aRAG-CoT (Kimi-K2.5)                        & 20.00 & 68.90 & 28.38 & 63.27 \\
& aRAG-CoT (Qwen3-14B)                        & 18.75 & 69.04 & 28.00 & 61.79 \\
& aRAG-CoT (DeepSeek-R1-Distill-Llama-8B)     & 35.00 & 81.54 & 30.31 & 66.25 \\
\cdashline{2-6}
& \textbf{\method} (GPT-5.2)                     & 47.89 & 81.31 & 32.13 & 68.34 \\
& \textbf{\method} (Gemini-3.1-Pro)              & \textbf{52.11} & \textbf{84.01} & 32.60 & 68.41 \\
& \textbf{\method} (Kimi-K2.5)                   & 44.93 & 82.72 & \underline{35.21} & \underline{72.44} \\
& \textbf{\method} (Qwen3-14B)                   & 43.66 & 81.96 & 33.60 & 69.44 \\
& \textbf{\method} (DeepSeek-R1-Distill-Llama-8B)& 45.07 & 82.50 & 31.52 & 70.11 \\
\bottomrule
\end{tabular*}

\end{table*}

\noindent\textbf{Baselines \& Implementation Details}. We select the following baselines for comparison, including nine conventional text classifiers: BioInfo~\citep{oliveira2020bioinfo}, iLab~\citep{iLab}, RELAI~\citep{relai}, Sense2Vec~\citep{perez2022automatic}, Voting~\citep{perez-etal-2023-semantic}, Recall~\citep{perez-etal-2023-semantic}, DUTH~\citep{spartalis-etal-2021-transfer}, Symanto~\citep{basile-etal-2021-upv-symanto}, and CYUT~\citep{wu-qiu-2021-roberta}, as well as one LLM-based method, aRAG-CoT~\citep{ravenda-etal-2025-llms}~\footnote{
The original CoT-based implementation of aRAG used Claude 3.7 Sonnet, whose official API is no longer available. We therefore implement the CoT setting using currently accessible models.
}  .  
For conventional methods, we follow their default experimental settings. For LLM-based methods, the reasoning effort is set to \texttt{medium}, and the maximum output length is set to \(8192\) tokens to prevent output truncation. For our \method, we split users' historical social traces into chunks of length 50. BGE-M3 is used as the embedding model for retrieval, and the top-25 candidate posts are retained for each scale item as memory. Our method is evaluated in a few-shot setting across all main experiments. For the LLM backbone, closed-source models and Kimi-K2.5 are accessed through their official APIs. For open-source models, including Qwen3-14B and DeepSeek-R1-Distill-Llama-8B, are locally deployed with the latest vLLM and tested on a single NVIDIA RTX A6000 GPU. We report the mean performance over three independent runs.

\begin{figure}[htb]
\vspace{-2mm}
    \centering
    \includegraphics[width=0.45\textwidth]{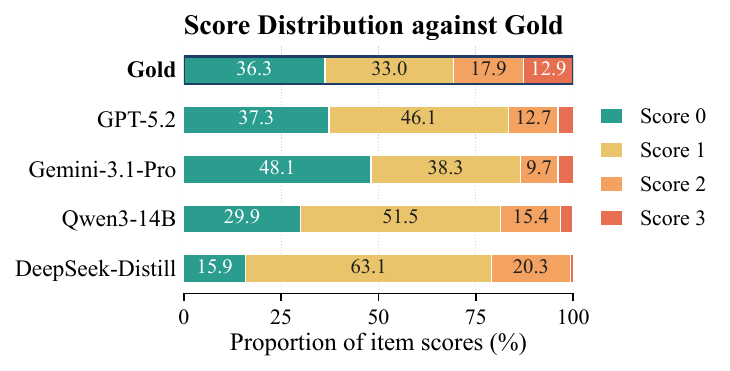}
    \caption{The score distribution between proprietary LLM and Open-weight LLM}
    \label{fig:score_distribution}
    \vspace{-2mm}
\end{figure}
\subsection{Performance on \task}

\noindent\textbf{Overall Performance}. The overall performance on \task~are shown in~\cref{tab:main_results}. It is evident that our \method~achieves competitive performance across both eRisk 2020 and eRisk 2021.
Specifically, \method~achieves the best performance on symptom-level metrics on the eRisk 2020 dataset and questionnaire-level metrics on the eRisk 2021 dataset. For the rest of the metrics, \method~still performs competitively, with results close to the SOTA methods. Notably, the strongest baseline Voting~\cite{perez-etal-2023-semantic} benefits from a voting strategy built upon additional manual annotations. In contrast, our \method~is training-free, operates under a few-shot setting, making it more adaptable to different psychological questionnaires (see cross-task results in~\cref{sec:exp-gen}). Beyond performance, \method~also provides stronger interpretability by grounding item-level scoring in retrieved evidence and structured reasoning.

\noindent\textbf{Impact of LLM backbones} We further analyze the impact of different LLM backbones on LLM-based methods, including aRAG and our \method, and obtain the following observations: (1) Compared with aRAG, \method~achieves better and more balanced performance across all metrics, even when equipped with a lightweight backbone. (2) Under our \method, open-source backbones obtain performance comparable to closed-source LLMs, while closed-source models show only slight advantages, possibly due to their stronger summarization and reasoning abilities. (3) Larger and smaller open-source backbones exhibit relatively comparable performance, indicating that \method~is not highly dependent on model scale. To further analyze the scoring behavior of different backbones, we visualize the estimated score distributions in~\cref{fig:score_distribution}.  Most models tend to favor low-severity responses, especially scores 0 and 1, while smaller models assign relatively more predictions to scores 1 and 2. This tendency may help lightweight models obtain competitive questionnaire-level results by producing aggregate severity estimates closer to the ground truth. However, their item-level metrics remain weaker than those of proprietary LLMs, suggesting that lightweight models still struggle with fine-grained rubric following and item-level score calibration.
In sum, these findings demonstrate the cross-LLM stability of \method, which mainly stems from its tailored modules: user-specific mental prior distillation provides user-level contextual information, scale-guided multi-path retrieval supplies more generalizable and reliable item-relevant evidence, and fine-grained evidence-aware reasoning guides the LLM to interpret the retrieved evidence more accurately.

\noindent\textbf{Human Evaluations}
Three annotators are asked to independently rate the extracted user-specific mental prior on a 1–5 scale based on the corresponding user-specific mental prior, assessing: (1) Objective Accuracy, namely whether the prior accurately reflects objective facts without introducing hallucinated information; (2) Completeness, namely whether it covers the important user-specific information without significant omissions; and (3) Subjective Reasonableness, namely whether subjective descriptions are rational, unbiased, and adequately supported by mental prior. For each metric, we consider the annotators to be in agreement when their ratings differ by no more than one point on the five-point scale. As shown in~\cref{tab:prior-quality-human-evaluation}, all three evaluation metrics receive relatively high average scores from the three annotators. Under this one-point tolerance criterion, the agreement rates among the three annotators are 86.3\% for Objective Accuracy, 73.1\% for Completeness, and 87.9\% for Subjective Reasonableness. These results suggest that the extracted mental priors can be evaluated with relatively consistent human judgments. 

\begin{table}[htb]
    \centering
    \caption{Average scores assigned by three human annotators for each mental prior quality metric.}
    \renewcommand{\arraystretch}{1.05}

    \resizebox{\columnwidth}{!}{%
        \begin{tabular}{@{}lcccc@{}}
            \toprule
            Dataset
            & Objective Accuracy
            & Completeness
            & Subjective Reasonableness \\
            \midrule
            eRisk-Avg
            & 4.53
            & 4.02
            & 4.14 \\
            \bottomrule
        \end{tabular}%
    }
    \label{tab:prior-quality-human-evaluation}
\end{table}

\begin{table}[htb]
\centering
\small
\vspace{-2mm}
\caption{Ablation study on eRisk 2020.}
\vspace{-2mm}
\label{tab:ablation}
\renewcommand{\arraystretch}{1.08}
\setlength{\tabcolsep}{3pt}
\resizebox{\columnwidth}{!}{%
\begin{tabular}{lccccc}
\toprule
\multirow{2}{*}{\textbf{Configuration}} 
& \multirow{2}{*}{\textbf{MAE} ($\downarrow$)}
& \multicolumn{2}{c}{\textbf{Symptom Metrics}}
& \multicolumn{2}{c}{\textbf{Questionnaire Metrics}} \\
\cmidrule(lr){3-4} \cmidrule(lr){5-6}
& & \textbf{AHR}~\% ($\uparrow$) & \textbf{ACR}~\% ($\uparrow$) 
& \textbf{ADODL}~\% ($\uparrow$) & \textbf{DCHR}~\% ($\uparrow$) \\
\midrule
\textbf{\method}    & \textbf{0.8082} & \textbf{41.97} & \textbf{73.77} & \textbf{84.13} & \textbf{40.00} \\
w/o $\mathcal{S}_u$ & 0.8707          & 38.57          & 70.97          & 81.72          & 28.57          \\
w/o SMR   & 0.8361          & 40.27          & 72.13          & 82.56          & 31.43          \\
w/o EaR    & 0.8721          & 39.12          & 70.93          & 81.68          & 38.57          \\
\bottomrule
\end{tabular}%
}
\vspace{-4mm}
\end{table}

\subsection{Ablation Study}
To analyze the contribution of each component, we conduct an ablation study on the eRisk 2020 dataset. We additionally report item-level Mean Absolute Error (MAE) to quantify the average absolute deviation between the predicted and gold BDI-II total scores~\citep{willmott2005advantages}. 
As shown in~\cref{tab:ablation}, we draw the following observations:
(\textbf{O1}) After removing the user-specific mental prior (w/o \(\mathcal{S}_u\)), the performance drops substantially across all metrics, demonstrating that user-level mental priors are crucial for user-perspective reasoning. This also supports our user-specific formulation, indicating that depression severity estimation requires not only retrieved posts, but also a holistic understanding of the user's inner characteristics. We further present a case study in Appendix~\ref{app:case_study}, where we find that our distilled user-specific mental prior provides a comprehensive summary of the user's long-term social characteristics, such as habitual behavioral patterns. This user-level prior helps the model understand the user's mental tendencies and self-expression style, enabling it to interpret indirect evidence from memories and produce judgments that are better aligned with the user's self-reported ground truth. 
%
(\textbf{O2}) Removing the EaR module (w/o EaR) also leads to clear performance degradation, highlighting that fine-grained evidence assessment is essential for reliable item-level reasoning. Without EaR, the model is more likely to directly predict scores from retrieved posts based on coarse semantic matching or surface-level keyword overlap, which weakens its ability to interpret implicit mental-state evidence. To further examine whether EaR effectively guides the model to infer indirect evidence by leveraging the user-specific mental prior, rather than simply relying on explicitly retrieved posts, we employ GPT-4o as an LLM-based judge for automatic evaluation. Specifically, we introduce \textit{Content Attention Analysis} (CAA) as the judge metric to assess whether the generated rationales mainly rely on the mental prior, retrieved posts, or their joint use. Due to space limitations, the detailed results and analysis are reported in Appendix~\ref{app:CAA}.
(\textbf{O3}) The ``w/o SMR'' variant is implemented by replacing SMR with a standard retrieval strategy. The results show consistent degradation across all evaluation metrics, with the most pronounced drop observed in DCHR. This finding indicates that the proposed scale-guided multi-query retrieval mechanism is effective in collecting more comprehensive item-relevant evidence from multiple semantic perspectives. By providing richer and more diverse evidence for simulating users' self-reported depression severity, SMR improves both the accuracy and consistency of the predicted scores. We further analyze the effects of different embedding encoders and top-\(K\) settings, with results reported in Appendix~\ref{app:Embedding_Model_replacement} and Appendix~\ref{app:retrieval_depth}, respectively.

\subsection{Cross-task Generalization}
\label{sec:exp-gen}
\begin{figure}[htb]
\vspace{-2mm}
    \centering
\includegraphics[width=0.46\textwidth]{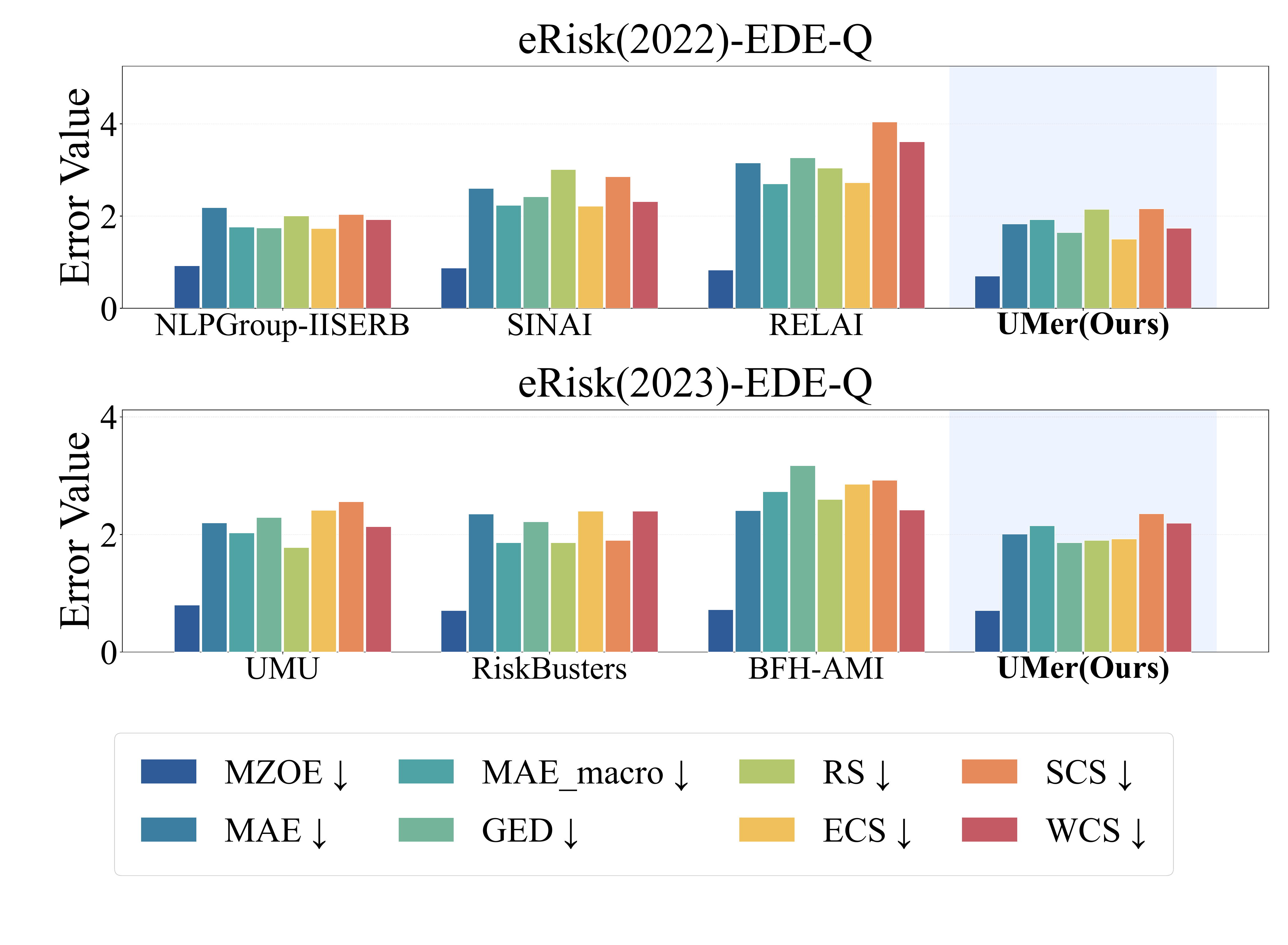}
    \caption{Generalization evaluation on eRisk 2022 and eRisk 2023 EDE-Q scoring tasks. \textbf{Lower values indicate better performance.}}
    \label{fig:transfer_experiment}
    \vspace{-8mm}
\end{figure}

We evaluate \method~on two additional questionnaire-based eating-disorder assessment tasks, i.e., the EDE-Q scoring tasks in eRisk 2022 and eRisk 2023. 
We select the following baselines for comparison: SINAI~\citep{marmolromero2022sinai} and RELAI~\citep{saravani2022measuring} employ training-free similarity-based strategies. NLPGroup-IISERB~\citep{srivastava2022nlpiiserb} leverages a BERT model fine-tuned on an external anorexia detection dataset. UMUTeam~\citep{pan2023umuteam}, RiskBusters~\citep{grigore2023transformer}, and BFH-AMI~\citep{merhbene2023bfhami} all train task-specific prediction components using labeled EDE-Q response data. 
Specifically, we use GPT-5.2 as the LLM backbone and keep all other settings unchanged. Since no auxiliary corpus is available for query diversification, we directly use the EDE-Q items and their explanations as retrieval queries. The user-specific mental prior construction and structured reasoning follow the same pipeline as in \task, and the detailed reasoning prompt template is provided in Appendix~\ref{app:reasoning_prompt}. 
The experimental results are shown in~\cref{fig:transfer_experiment}, where we observe that our \method~achieves balanced and competitive performance on both eRisk 2022 and eRisk 2023 EDE-Q scoring tasks.
Notably, compared with the task-specific baselines, our training-free framework achieves competitive or even superior performance across several metrics. These results suggest that the proposed user-centered, mental-guided self-reported framework can be effectively transferred to another continuous psychological scale, demonstrating strong generalization ability under zero-training or few-shot settings.

\section{Conclusion}
In this paper, we formalize social traces-based depression severity estimation as a user-centered self-report simulation task. Instead of directly diagnosing users based on their social media posts, our goal is to use these posts as evidence to simulate how users would report their own psychological states when completing standardized scale items, thereby making the predicted scores better aligned with users' self-reported ground truth and enabling a more faithful estimation of their psychological status.
To this end, we propose \method, a user-centered mental evidence reasoning framework that combines user mental prior construction for capturing user-level psychological priors, scale-aware evidence retrieval for collecting item-relevant posts from multiple semantic perspectives, and evidence-aware structured reasoning for distinguishing direct and indirect evidence and performing item-level scoring.
Extensive experiments on real-world benchmarks demonstrate that \method~achieves competitive performance, provides evidence-grounded interpretability, and generalizes to other questionnaire-based mental health assessment tasks under zero-training or few-shot settings. 
Limitations and ethical considerations are discussed in the supplementary material.
%

\section{Limitations}

Despite the effectiveness of our proposed \method, several limitations remain:

\noindent(1) \textbf{Cost and Data Dependency of User Mental Prior}. 
\method~requires distilling a user mental prior from historical social traces. On the one hand, this introduces additional computational cost compared with existing LLM-based methods, since constructing a user-level prior from scratch for each user can be expensive. On the other hand, the recursive updating strategy is more suitable for long-term user modeling scenarios. Therefore, sufficient historical traces are indispensable for reliable prior construction. 
Future work should explore simpler and more efficient ways to extract user-level priors, such as lightweight profile summarization or retrieval-based approximation.

\noindent (2) \textbf{Dataset Scale and Severity Calibration}. 
The utilized datasets contain a relatively small number of users, and the overall severity distribution is skewed toward moderate cases. This may make the model more prone to underestimating severe conditions, especially when strong evidence is scarce or implicitly expressed in social media posts. More robust calibration strategies are needed to reduce systematic underestimation and improve prediction reliability across different severity levels.

\noindent (3) \textbf{Clinical Use and Interpretation}.  The proposed \method~is developed to assist preliminary screening and questionnaire-based research, rather than to provide definitive medical diagnosis or clinical advice. Although \method~simulates users' self-reporting process to estimate their psychological states with standardized psychological scales, its outputs should be interpreted as auxiliary signals and should not replace professional assessment by qualified mental health practitioners.

\section{Ethical Considerations}

This work uses social media data for questionnaire-based mental health assessment, which requires careful attention to privacy, data protection, and the boundaries of system use. Users' posts may contain sensitive personal information, and the generated mental prior should also be regarded as sensitive derived data. Therefore, any deployment of such a system should follow strict data governance principles, including informed consent when applicable, data minimization, anonymization, secure storage, and restricted access.

The proposed framework is intended for preliminary screening and research-oriented questionnaire completion, rather than definitive clinical diagnosis or medical decision-making. Although standardized psychological scales provide a structured way to estimate users' self-reported psychological states, model predictions may still be affected by incomplete evidence, biased expressions, and reasoning errors. The outputs should therefore be interpreted as auxiliary signals for identifying potential risk and should not replace assessment by qualified mental health professionals.

Responsible use also requires preventing over-interpretation of model-generated scores and explanations. The system should not be used in high-stakes settings without human oversight, clinical validation, and appropriate safeguards. Users should be clearly informed that the framework simulates questionnaire responses from historical posts, instead of directly diagnosing their mental health status.

\bibliography{custom}

\clearpage
\appendix
\renewcommand\thefigure{A\arabic{figure}}
\renewcommand\thetable{A\arabic{table}}  
\renewcommand\theequation{A\arabic{equation}}
\setcounter{equation}{0}
\setcounter{table}{0}
\setcounter{figure}{0}

\section*{Appendix}

\section{Additional Experimental Results}


\subsection{Case Study For Mental Prior}
\label{app:case_study}
\begin{figure*}[htb]
    \centering
    \includegraphics[width=0.9\textwidth]{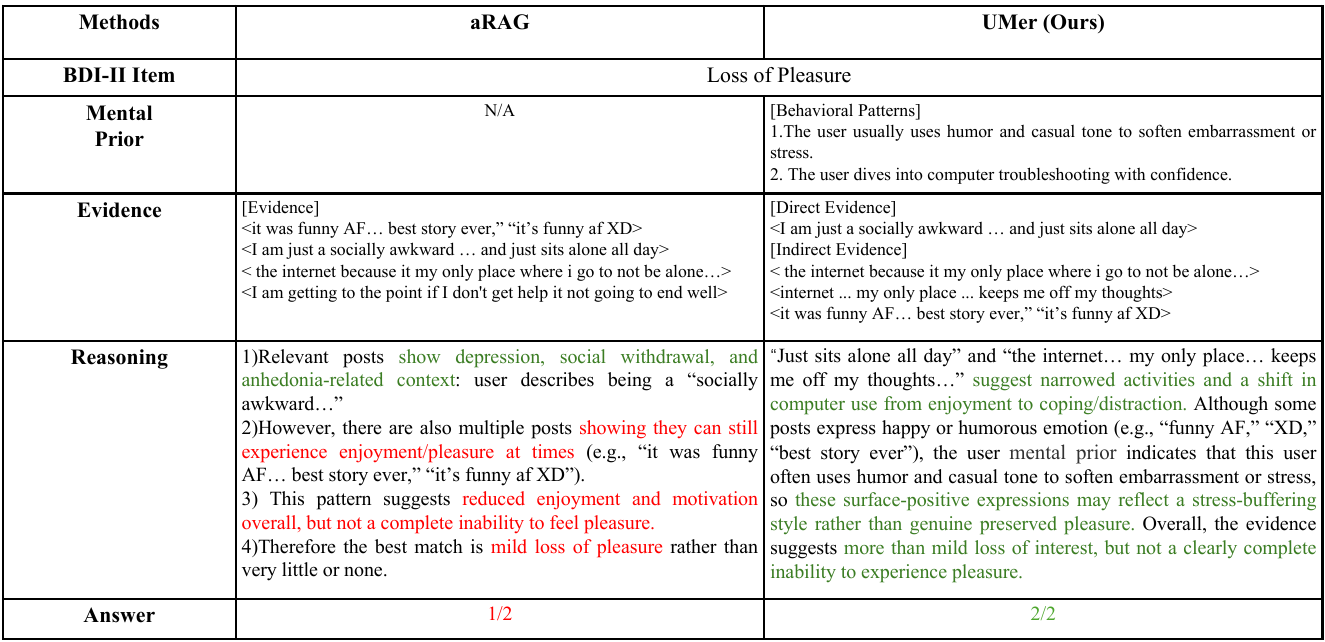}
    \caption{Case study for mental prior}
    \label{fig:case_study_persona}
    \vspace{-4mm}
\end{figure*}

As shown in~\cref{fig:case_study_persona}, case study showing the effect of the user-specific mental prior. Compared with aRAG, the distill prior provides rich user-level context about mental tendencies, behavioral patterns, and self-expression style, enabling \method~to better interpret indirect evidence and produce predictions more aligned with the ground truth.

\subsection{The Content Attention Analysis} 
\label{app:CAA}
\begin{figure}[htb]
\vspace{-2mm}
    \centering
    \includegraphics[width=0.48\textwidth]{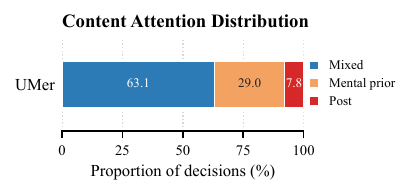}
    \vspace{-6mm}
   \caption{Content attention analysis. "Mixed" denotes reasoning that jointly relies on information from both the mental prior and retrieved posts.}
    \label{fig:llm_judge}
    \vspace{-4mm}
\end{figure}

\begin{figure*}[!t]
    \centering
    \includegraphics[width=0.92\textwidth]{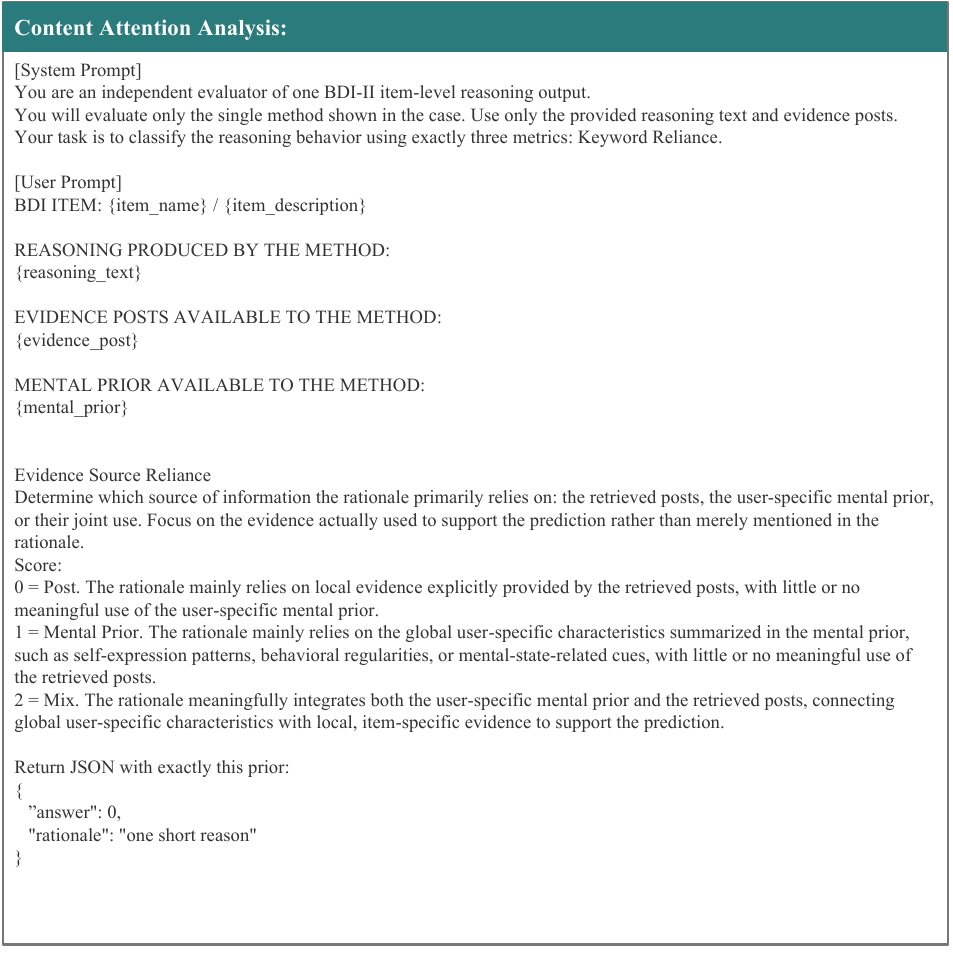}
    \caption{Prompt template for key-LLM-based query generation.}
    \label{fig:caa_llm_prompt}
\end{figure*}

We further employ GPT-4o as an LLM-based judge to analyze the information sources emphasized in the generated rationales, which we refer to as \textit{Content Attention Analysis}. Specifically, the judge classifies each rationale according to whether its reasoning primarily focuses on the retrieved posts (\textit{Post}), the user-specific mental prior (\textit{Mental Prior}), or an integrated use of both sources (\textit{Mixed}). The evaluation metric used by the LLM judge are provided in~\cref{fig:caa_llm_prompt}.As shown in~\cref{fig:llm_judge}, \method~more frequently generates rationales by jointly integrating the global user-specific information captured by the mental prior with the local evidence provided by individual posts. Even when the rationale relies primarily on a single evidence source, \method~shows a tendency to incorporate the mental prior rather than only the retrieved posts. This result indicates that the mental prior can help to capture informative self-expression patterns, behavioral regularities, and mental-state-related cues that complement sparse post-level evidence. Moreover, the fine-grained reasoning procedure guides the LLM to appropriately connect these summarized characteristics with item-specific evidence, thereby supporting more reliable psychological scale prediction.

\subsection{Impact of the Embedding Encoder}
\label{app:Embedding_Model_replacement}
We conduct additional experiments on the eRisk 2020 by replacing the default embedding encoder $\phi$ used in the embedding retriever while keeping the other components unchanged. The results are reported in~\cref{tab:retriever_comparison}. 
We find that BGE-M3 achieves the best results across almost all metrics, while the GTE multilingual base obtains the highest DCHR. Importantly, replacing BGE-M3 with other embedding models does not cause a severe performance degradation across the evaluation metrics. This result suggests that the proposed multi-query retrieval strategy provides relatively stable evidence coverage across different retriever backbones.

\begin{table}[htb]
\centering
\caption{Performance of different embedding encoders.}
\label{tab:retriever_comparison}
\renewcommand{\arraystretch}{1.08}
\setlength{\tabcolsep}{3pt}
\resizebox{\columnwidth}{!}{%
\begin{tabular}{lccccc}
\toprule
\multirow{2}{*}{\textbf{Retriever}} 
& \multirow{2}{*}{\textbf{MAE} $\downarrow$}
& \multicolumn{2}{c}{\textbf{Symptom Metrics}}
& \multicolumn{2}{c}{\textbf{Questionnaire Metrics}} \\
\cmidrule(lr){3-4} \cmidrule(lr){5-6}
& & \textbf{AHR}~\% ($\uparrow$) & \textbf{ACR}~\% ($\uparrow$) 
& \textbf{ADODL}~\% ($\uparrow$) & \textbf{DCHR}~\% ($\uparrow$) \\
\midrule
\textbf{BGE-M3}  & \textbf{0.8082} & \textbf{41.97} & \textbf{73.77} & \textbf{84.13} & 40.00 \\
GTE multilingual base   & 0.8354          & 38.91          & 72.15          & 83.13          & \textbf{44.29} \\
Qwen3 embedding 0.6B   & 0.8517          & 39.66          & 71.61          & 82.18          & 38.29 \\
\bottomrule
\end{tabular}%
}
\end{table}

\subsection{Impact of Retrieval Depth}
\label{app:retrieval_depth}
We investigate the influence of retrieval depth by varying the number of retrieved posts used for mental memory construction. The experiments are conducted on eRisk 2020, and the results are shown in~\cref{tab:retrieval_depth}. We find that retrieving the top-25 posts achieves the best performance across all reported metrics. Retrieving fewer posts may omit relevant evidence, whereas retrieving more posts may introduce noisy or irrelevant content, thereby interfering with evidence-grounded judgment.

\begin{table}[htb]
\centering
\caption{Performance of different retrieval depths. }
\label{tab:retrieval_depth}
\renewcommand{\arraystretch}{1.08}
\setlength{\tabcolsep}{3pt}
\resizebox{\columnwidth}{!}{%
\begin{tabular}{lccccc}
\toprule
\multirow{2}{*}{\textbf{Configuration}} 
& \multirow{2}{*}{\textbf{MAE} $\downarrow$}
& \multicolumn{2}{c}{\textbf{Symptom Metrics}}
& \multicolumn{2}{c}{\textbf{Questionnaire Metrics}} \\
\cmidrule(lr){3-4} \cmidrule(lr){5-6}
& & \textbf{AHR}~\% ($\uparrow$) & \textbf{ACR}~\% ($\uparrow$) 
& \textbf{ADODL}~\% ($\uparrow$) & \textbf{DCHR}~\% ($\uparrow$) \\
\midrule
\textbf{top-25} & \textbf{0.8082} & \textbf{41.97} & \textbf{73.77} & \textbf{84.13} & \textbf{40.00} \\
top-10          & 0.8422          & 39.66          & 71.93          & 82.22          & 30.00          \\
top-15          & 0.8395          & 39.46          & 72.02          & 82.45          & 28.57          \\
top-20          & 0.8388          & 39.18          & 72.04          & 82.93          & 32.86          \\
top-30          & 0.8279          & 39.93          & 72.40          & 82.84          & 34.29          \\
top-35          & 0.8401          & 39.32          & 72.00          & 82.38          & 30.00          \\
top-40          & 0.8406          & 39.27          & 71.98          & 82.75          & 33.33          \\
\bottomrule
\end{tabular}%
}
\end{table}

\section{Prompt Templates}
\label{app:prompt}

\subsection{Mental prior distilling skill}
\label{app:ms_prompt}
\begin{figure*}[!t]
    \centering
    \includegraphics[width=0.92\textwidth]{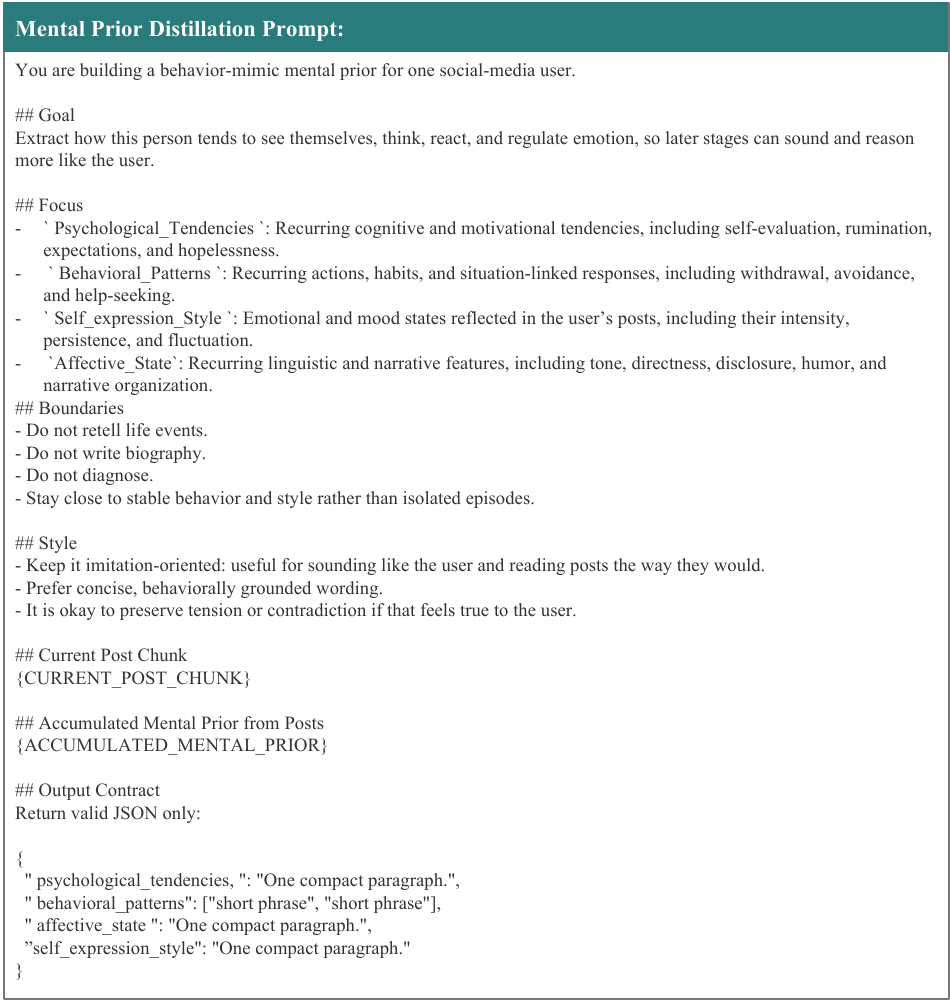}
    \caption{Prompt template for constructing the accumulated mental prior from posts.}
    \label{fig:ms_prompt}
\end{figure*}

In~\cref{fig:ms_prompt}, we present the prompt used to recursively construct the user mental prior from the user's prior social traces. Specifically, for each social trace chunk, the profiling skill guides the LLM to act as an indirect behavioral analyst and extract only evidence-supported, self-report-relevant patterns, rather than making clinical diagnoses. The skill produces a structured prior containing four fields: \textit{affective and linguistic expression style}, \textit{symptom disclosure tendency}, \textit{self-attribution style} and \textit{behavioral regularities}. Each field is supported by brief evidence from the current chunk. The chunk-level priors are then compressed into a global user mental prior by retaining stable, repeated, and scale-relevant patterns while removing redundant or one-off observations.

\subsection{Multi Query Diversification}
\label{app:mqc_prompt}
\begin{figure*}[!t]
    \centering
    \includegraphics[width=0.92\textwidth]{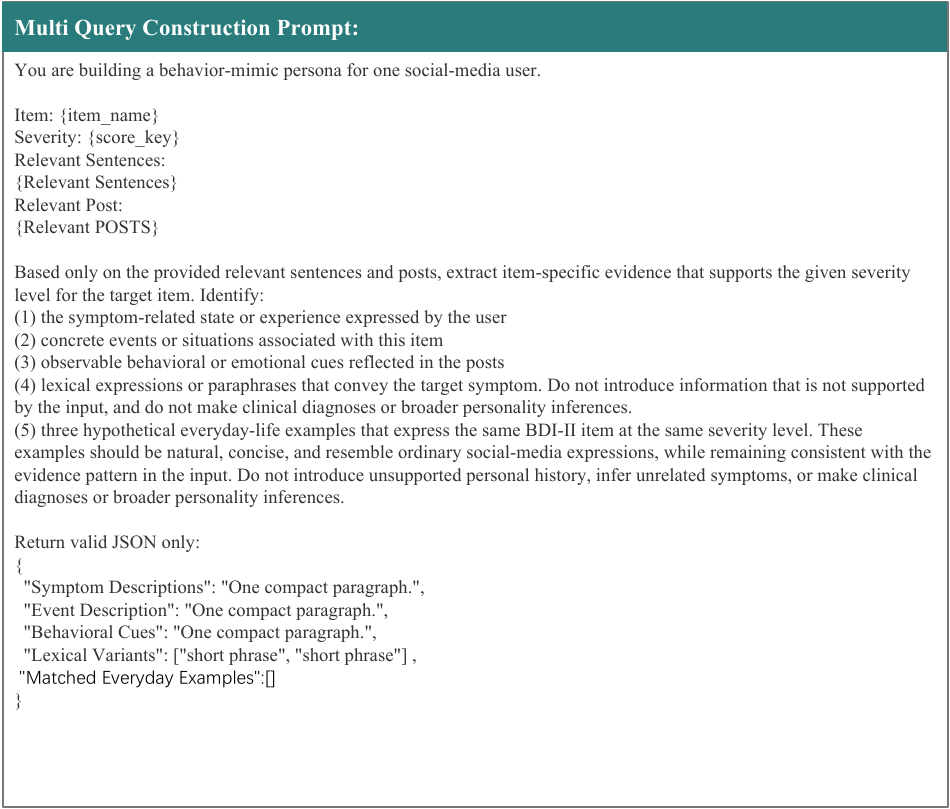}
    \caption{Prompt template for multi-query construction.}
    \label{fig:mqc_prompt}
\end{figure*}

To bridge the semantic gap between standardized scale item definitions and informal social media expressions, we use a prompt to diversify retrieval queries for each scale item from multiple perspectives. 
Given the item description and rubric examples, the prompt generates diverse evidence-seeking queries that cover explicit symptom expressions, implicit behavioral or psychological cues, and everyday narrative variants. Each query path independently retrieves candidate posts, which are then merged using the max-score fusion strategy described in the main text. The prompt template for scale-guided multi-query construction is shown in~\cref{fig:mqc_prompt}.

\subsection{Evidence-aware Reasoning}
\label{app:reasoning_prompt}
\begin{figure*}[!t]
    \centering
    \includegraphics[width=0.92\textwidth]{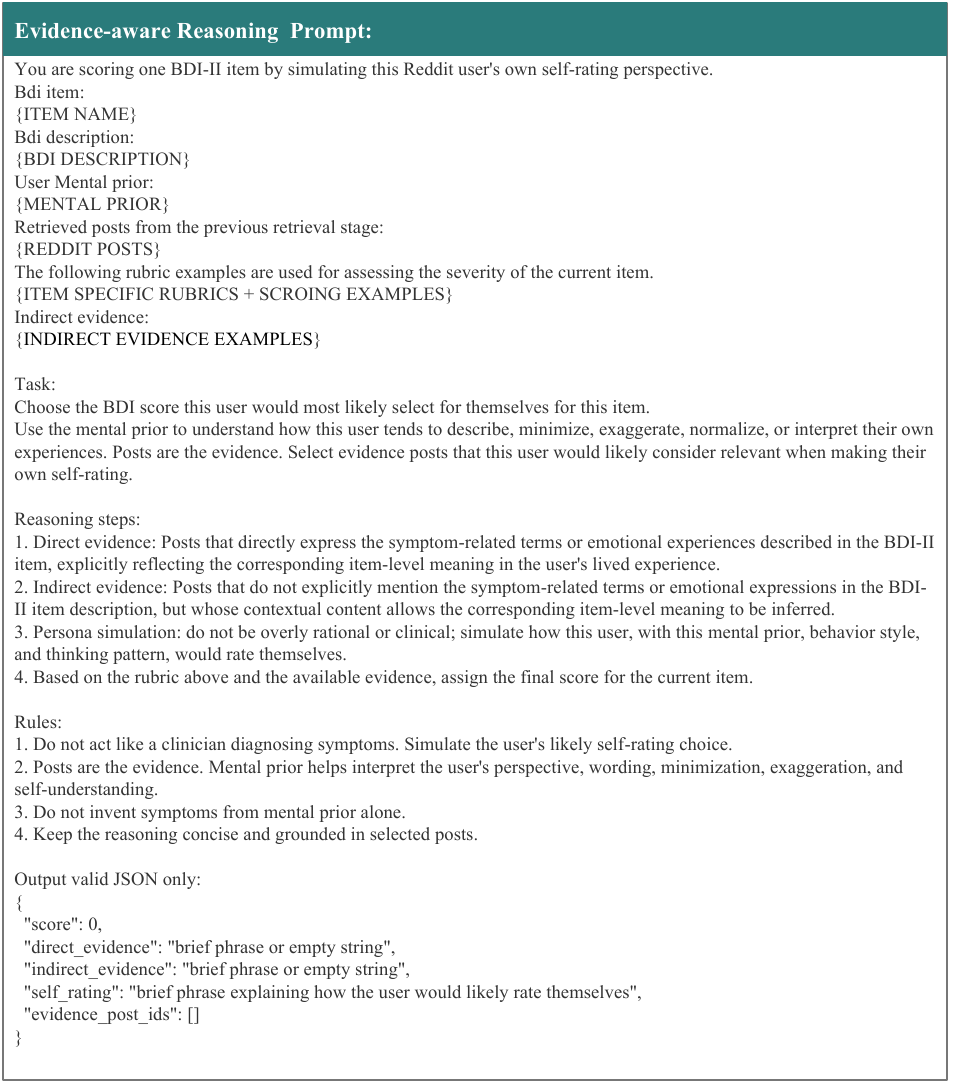}
    \caption{Prompt template for structured reasoning and final item-level self-rating simulation.}
    \label{fig:reasoning_prompt}
\end{figure*}

The prompt template for evidence-aware reasoning and final item-level prediction is shown in~\cref{fig:reasoning_prompt}. Specifically, for each scale item, the judgment prompt integrates the item description, the diversified rubric set, the retrieved mental evidence memories, and the distilled user mental prior. It guides the model to examine whether the retrieved posts provide direct or indirect evidence for the target item, assess how strongly the evidence supports different score levels, and finally produce an item-level self-rating with a brief evidence-grounded rationale. This design supports interpretable questionnaire completion without requiring the model to output unrestricted reasoning traces.

\subsection{Cross-task Generalization Prompt}
\label{app:gae}
\begin{figure*}[!t]
    \centering
    \includegraphics[width=0.92\textwidth]{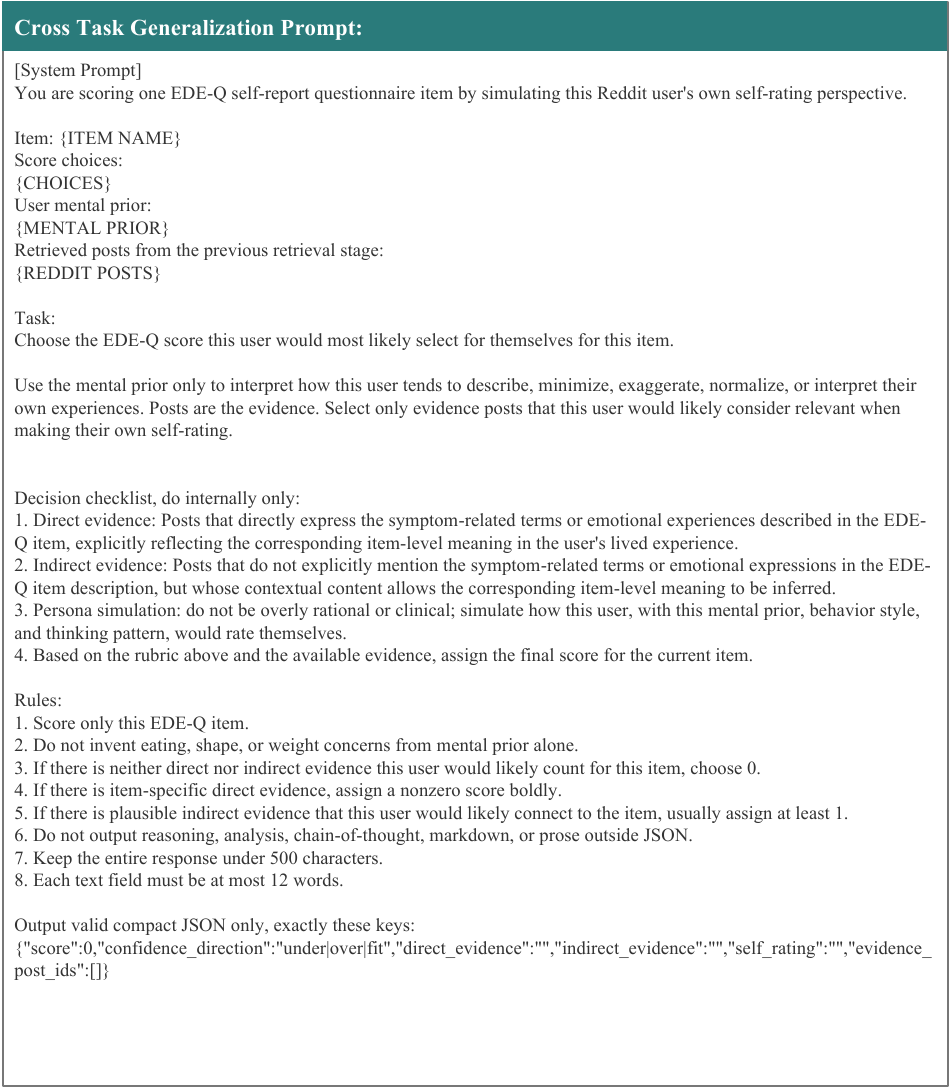}
    \caption{Prompt template for Generalization ability Experiment.}
    \label{fig:gae}
\end{figure*}

For the cross-task generalization experiment, we adapt our prompting framework to the eRisk eating disorder assessment task by simply adjusting the scale-related information. Specifically, given the target EDE-Q questionnaire, \method~predicts item-level scores and supporting evidence. The complete prompt used in the generalization experiment is presented in~\cref{fig:gae}.

\end{document}